%% file: arxiv.tex
\documentclass[10pt,twocolumn,letterpaper]{article}

\newif\ifarxiv
\arxivtrue

\ifarxiv    \usepackage[pagenumbers]{wacv} 
\else       \usepackage{wacv}      
\fi

\usepackage{graphicx}
\usepackage{amsmath}
\usepackage{amssymb}
\usepackage{booktabs}
\usepackage{multirow}
\usepackage{xcolor}
\usepackage{arydshln}
\usepackage[symbol]{footmisc}
\usepackage[accsupp]{axessibility}  

\makeatletter
\def\adl@drawiv#1#2#3{%
        \hskip.5\tabcolsep
        \xleaders#3{#2.5\@tempdimb #1{1}#2.5\@tempdimb}%
                #2\z@ plus1fil minus1fil\relax
        \hskip.5\tabcolsep}
\newcommand{\cdashlinelr}[1]{%
  \noalign{\vskip\aboverulesep
           \global\let\@dashdrawstore\adl@draw
           \global\let\adl@draw\adl@drawiv}
  \cdashline{#1}
  \noalign{\global\let\adl@draw\@dashdrawstore
           \vskip\belowrulesep}}
\makeatother

\usepackage{subcaption}
\usepackage{makecell}
\usepackage{cuted}
\usepackage{colortbl}

\usepackage[pagebackref,breaklinks,colorlinks]{hyperref}

\usepackage[capitalize]{cleveref}
\crefname{section}{Sec.}{Secs.}
\Crefname{section}{Section}{Sections}
\Crefname{table}{Table}{Tables}
\crefname{table}{Tab.}{Tabs.}

\begin{document}

\title{RADIO: Reference-Agnostic Dubbing Video Synthesis}

\author{Dongyeun Lee$^{1,}$\thanks{Equal Contribution. $^\dagger$ Corresponding Authors.} \quad Chaewon Kim$^{1,}$\footnotemark[1] \quad Sangjoon Yu$^{1}$ \quad Jaejun Yoo$^{2,\dagger}$ \quad  Gyeong-Moon Park$^{3,\dagger}$ \\ 
$^{1}$Klleon AI Research \qquad $^{2}$UNIST \qquad $^{3}$Kyung Hee University
}

\twocolumn[{%
\vspace{-2mm}
\renewcommand\twocolumn[1][]{#1}%
\maketitle
\vspace{-7mm}
\begin{center}
    \centering
    \begin{tabular}{c}
        \begin{minipage}{0.96\textwidth}
        \includegraphics[width=\linewidth]{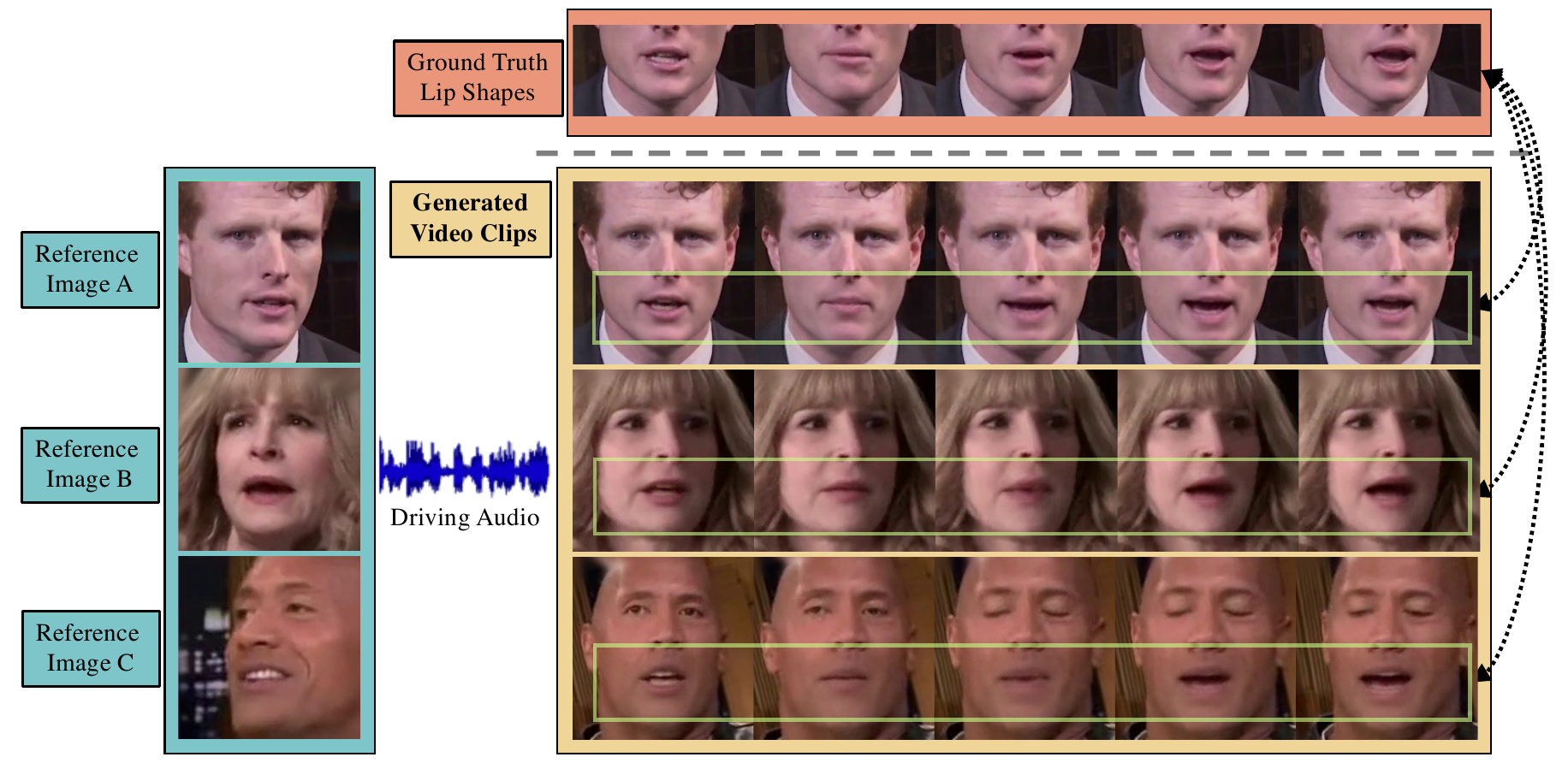}
        \end{minipage}
\hspace{-0.4cm}
\vspace{-0.1cm}
\\

\end{tabular}
\captionof{figure}{\textbf{
Illustration of results generated by our RADIO framework.} Our method targets a one-shot audio-driven talking face generation, where synchronized mouth shapes are generated while holding on to the identity of a single reference frame. Even with diverse poses and expressions of reference frames, our method generates accurately synced lips robustly.
}
\vspace{-0.1cm}
\end{center}%
}]


\input{section/0_abstract}
\input{section/1_introduction}

\input{section/2_rw}

\input{section/3_method}
\input{section/4_experiment}

\input{section/5_conclusion}

\input{section/6_acknowledgements}


{\small
\bibliographystyle{ieee_fullname}
\bibliography{egbib}
}

\ifarxiv
\clearpage
\appendix
\input{section/X_appendix}
\fi

\end{document}

%% file: section/0_abstract.tex

\begin{abstract}
\vspace{-0.14cm}

One of the most challenging problems in audio-driven talking head generation is achieving high-fidelity detail while ensuring precise synchronization. 
Given only a single reference image, extracting meaningful identity attributes becomes even more challenging, often causing the network to mirror the facial and lip structures too closely. 
To address these issues, we introduce RADIO, a framework engineered to yield high-quality dubbed videos regardless of the pose or expression in reference images. 
The key is to modulate the decoder layers using latent space composed of audio and reference features. Additionally, we incorporate ViT blocks into the decoder to emphasize high-fidelity details, especially in the lip region. 
Our experimental results demonstrate that RADIO displays high synchronization without the loss of fidelity. Especially in harsh scenarios where the reference frame deviates significantly from the ground truth, our method outperforms state-of-the-art methods, highlighting its robustness. 

\end{abstract}

%% file: section/1_introduction.tex
\section{Introduction}

Talking head generation~\cite{wang2021osfv, ren2021pirenderer, burkov2020latentpose, doukas2021headgan, zakharov2019fewshot} has become a focal point of research attention owing to its wide-ranging applications in the media industry, \eg virtual human animation, audio-visual dubbing, and video content creation. 
Audio-driven talking face generation specifically aims to produce high-quality videos that exhibit precise synchronization with the driving audio.
In particular, one-shot audio-driven methods are designed to generate talking faces of unseen speakers, given a single reference image. 

However, it is challenging to consistently generate high-quality synced faces, due to the risk of over-fitting to the single image. In other words, previous methods face difficulties to generate mouth shapes and poses that deviate from the source image. We observed that this problem can be attributed to the susceptibility of previously proposed frameworks to the choice of reference frame.
Early methods directly incorporate the information of reference image into the generator through skip-connections \cite{Parjwal2020wav2lip, kr2019towards, chung2017said, zhou2019disentangled}. These approaches constrain generated images to rarely diverge from the input image. Deformation-based methods \cite{chen2019hierarchical, zhang2021hdtf, zhou2020makeittalk, wang2022one, zhong2023iplap, zhang2023dinet} aim to adjust facial alignment based on audio or target frames, but still struggles to generate realistic diverse poses. Significant changes in geometric priors like mesh and landmarks, or latent space also introduce artifacts and distortions in the images 
\cite{suwaj2017obama, zhang2023dinet, yin2022styleheat, tang2022memface, zhou2021pcavs}.

Despite various efforts made by previous works, we observed that there has been a limited exploration into scenarios where the dubbed video demands significantly different generated frames compared to the reference image in terms of pose and mouth expression. In fact, this scenario is the most frequently encountered in reality, as it is both inconvenient and impractical to manually select the appropriate reference frame. The key to consistently produce high-quality dubbed video is to effectively and distinctly extract the identity-related characteristics while eliminating undesired elements such as pose, facial expression, and mouth shape from the reference image.

To address this issue, we introduce \textbf{RADIO} - a method for \textbf{R}eference-\textbf{A}gnostic \textbf{D}ubbing v\textbf{I}de\textbf{O} generation. RADIO aims to preserve high-fidelity details from the reference image and reduce sensitivity to the choice of the reference image, all within a unified framework.
Specifically, we adopt the decoder structure from StyleGAN2~\cite{Karras2019stylegan2}, and inject the reference frame information, \ie style feature,  through style-modulated convolution. 
Unlike previous methods with style-based generators as backbone, we do not inject the reference frame directly to the StyleGAN2 input \cite{zhou2021pcavs,yin2022styleheat,alghamdi2022talking-head}. 
While style modulation proves to be efficient in capturing identity-related features and diminishing structural reliance, it falls short in preserving high-fidelity details. 
To capture the fine texture and characteristic details of the source image, we introduce Vision Transformer (ViT) \cite{dosovitskiy2020vit} within the intermediate decoder layers, preceding the style modulation. 
With this simple framework, we can produce talking faces in a more practical and challenging scenario where the reference face significantly differs from the target face.


Our main contributions are summarized as follows:
\begin{itemize}
 \item We propose a simple yet effective architecture that extracts relevant information from a single reference image, thus able to create dubbing videos with improved lip synchronization that is robust from the reference pose or mouth shape.
 \item We improve fidelity preservation by incorporating carefully designed vision transformer blocks in the decoder, which specifically focus on lip-oriented details.
 \item We thoroughly evaluate RADIO with qualitative and quantitative experiments, and demonstrate its superiority over existing state-of-the-art methods.
\end{itemize}

%% file: section/2_rw.tex
\section{Related Works}

\subsection{Audio-Driven Talking Head Generation}

The task of audio-driven talking head generation learns to synthesize talking faces with lip movements synchronized with the driving audio. 
Early 3D-structure-based methods animate faces with 3D models such as meshes or vertex coordinates \cite{karras2017vertex, sarah2017speech, zhou2018viseme}. 
Unfortunately, the requirement of 3D model training data for individuals limits its application to animating general faces and struggles to reproduce teeth and hair details.
In response to this challenge, recent research has shifted towards directly animating raw 2D images. 2D-based audio-driven works comprehensively fall into two categories: generating talking faces in a speaker-specific or a speaker-agnostic manner.


\noindent\textbf{Speaker-specific methods.} Personalized models generate faces in a speaker-specific manner and require re-training for an unseen identity \cite{suwaj2017obama,thies2020nvp,yi2020audio,wen2020audiodvp,lu2021live,lahiri2021lipsync3d,guo2021adnerf,liu2022semanticaware,yao2022dfanerf,shen2022dfrf,li2023efficient}. Inspired by the development of neural rendering, recent methods model facial details implicitly by the hidden space of the neural radiance fields~\cite{mildenhall2020nerf}. AD-NeRF~\cite{guo2021adnerf} first proposes end-to-end audio-driven neural radiance fields for talking head generation. SSP-NeRF~\cite{liu2022semanticaware} introduces a semantic-aware dynamic ray sampling module, and DFA-NeRF~\cite{yao2022dfanerf} introduces two disentangled
representations for improvement of realistic dynamics. DFRF~\cite{shen2022dfrf} reduces the training speed via conditioning the face radiance field on 2D appearance images.
Nevertheless, the need for additional training efforts and the capability of NeRF to generalize to unfamiliar identities heavily restricts its applicability.

\input{fig/fig_architecture}

\noindent\textbf{Speaker-agnostic methods.}
Speaker-agnostic methods have gained popularity because they only require a single image of the target identity to animate the face with driving audio. Methods that generate the whole head either utilize warping techniques to drive the entire head movements \cite{chen2019hierarchical,zhou2020makeittalk,ji2021evp,zhou2021pcavs,wang2022one,zhang2021hdtf,Liang_2022_CVPR,ji2022eamm,zhang2023sadtalker}, or generate inverted images via a well-trained encoder and a pre-trained face generator\cite{yin2022styleheat,min2022styletalker,alghamdi2022talking-head}. The former approach has controllability over head motions; however, it comes at the expense of fidelity degradation and artifacts due to the shifting of facial landmarks. The latter approach produces high-quality images but carries the risk of generating images biased by the pre-trained generators, leading to the potential leakage of the identity information.

Methods that focus on mouth regions generate synchronized lip movements with the pose fixed by the target image. Inpainting-based methods~\cite{Parjwal2020wav2lip, park2022synctalkface, zhang2023dinet,  zhong2023iplap, guan2023stylesync} exhibit high accuracy in synchronization and identity preservation. However, in a one-shot scenario where only a single reference image is available, these models fail to preserve lip-oriented high-fidelity details. Furthermore, in harsh cases where the pose or expression of the reference image is significantly dissimilar to the target image, previous methods fail to robustly generate accurate mouth shapes. 
Our method focuses on extracting the high-fidelity identity information robustly from a single image, 
without the guidance of additional geometric face priors.

\subsection{Vision Transformer}

The significant success of transformers \cite{vaswani2017attention,brown2020language} in NLP has motivated numerous endeavors to extend their application to various vision tasks. 
Among them, Vision Transformer (ViT) \cite{dosovitskiy2020vit} has shown remarkable performance across several discriminative tasks\cite{liu2021swin,liu2022swin,li2022mvitv2,misra2021end,touvron2021training,cao2022swin,strudel2021segmenter,wang2021pyramid,dai2021up,xie2021segformer,zheng2021rethinking,kirillov2023segment}.
Concurrently, recent efforts have also emerged to explore the integration of ViT into generative tasks.
Several studies \cite{lee2021vitgan,zhang2022styleswin,zhao2021improved} have shown the competitive nature of ViT-based architectures when compared to CNN-based architectures \cite{karras2019style,Karras2019stylegan2,karras2021alias} as the unconditional generator.
Additionally, there have been attempts to utilize ViT in image-to-image translation \cite{kim2022instaformer,torbunov2023uvcgan}.
InstaFormer \cite{kim2022instaformer} leveraged ViT to capture the global consensus of a scene.
UVCGAN \cite{torbunov2023uvcgan} utilized ViT to learn pairwise relationships of low-frequency features.
They commonly incorporate self-attention modules at low-resolution layers to discover the global information from a given image.
On the other hand, our approach adopts ViT to generate high-fidelity results by capturing global relationships across features from different images in high-resolution layers.

%% file: fig/fig_architecture.tex
\begin{figure*}[ht!]
\begin{center}
\includegraphics[width=0.95\textwidth]{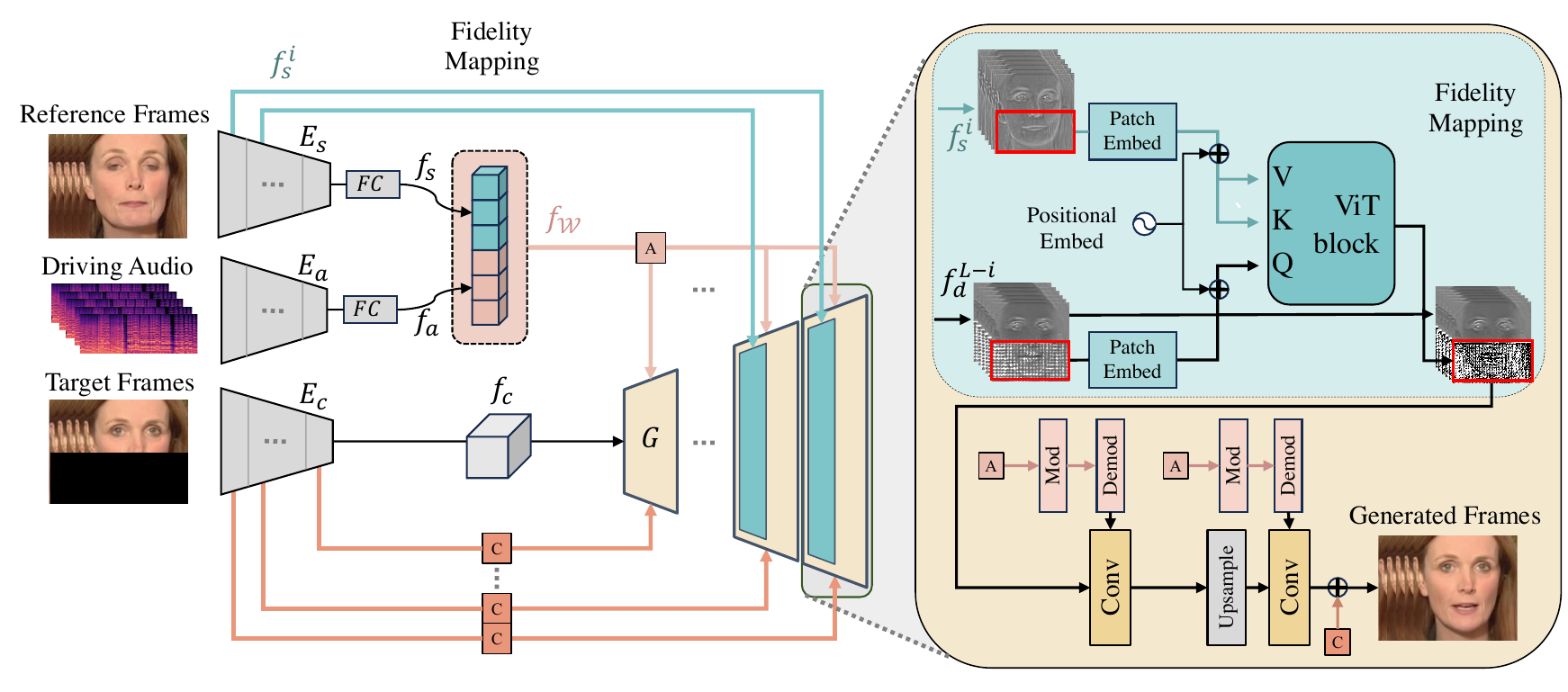}
\end{center}
\caption{
\textbf{Architecture of RADIO.} Our framework is composed of residual block encoders and a StyleGAN-based decoder. Lower half-masked target frames, reference frames, and mel-spectrograms are encoded by $E_c$, $E_s$, and $E_a$, respectively. Basically, the generator $G$ follows style modulation of StyleGAN2. The content feature $f_c$ is fed to the generator $G$ with residual mapping of the intermediate content features from each block of $E_c$. $f_\mathcal{W}$ is a concatenation of features $f_s$ and $f_a$, and is mapped to each generator block as a $\mathcal{W}$ space for style modulation. Inserted into the last two blocks of the generator, the ViT block receives the lower half of the $f_i$(query), an intermediate feature flowing through the generator layers, and the lower half of the $f_s^i$ (key, value), extracted from the front end of the style encoder $E_s$, with patch embedding and positional embedding. Finally, the output of ViT is re-concatenated with the upper half of the $f_i$ and passes through the aforementioned style modulation layer. The final frames are the audio-driven high-fidelity results.
}
\label{fig:architecture}
\end{figure*}

%% file: section/3_method.tex
\section{Method}

In this section, we propose RADIO, an efficient one-shot audio-driven talking face architecture. 
Figure \ref{fig:architecture} shows the overview of the architecture design. 
During the training phase, RADIO receives consecutive series of target frames $I_t \in \mathbb{R}^{T \times 3 \times H \times W}$, randomly chosen reference frames $I_r \in \mathbb{R}^{T \times 3 \times H \times W}$ of a target speaker, and an audio clip $A$ aligned with the corresponding $T$ frames as input. 
Our framework can create high-quality talking head videos $I_{out} \in \mathbb{R}^{T \times 3 \times H \times W}$ where the target face speaks with high synchronization agnostic to the facial alignment of reference frame. $T$ is the clip length and set to five, following the training strategy in \cite{Parjwal2020wav2lip} for the usage of sync discriminator \cite{Chung2016syncnet}. 
Note that in the inference phase, only a single target and reference frame are used as inputs, \ie $T$=1. 
$H$ and $W$ are the height and width of the frames, respectively.

\subsection{Notation and Proposed Architecture} \label{sec:architecture}

The proposed framework consists of four components: (\lowercase{\romannumeral1}) a content encoder $E_c$ for extracting the structural details of the target image, (\lowercase{\romannumeral2}) a style encoder $E_s$ for capturing the visual attributes linked to the target identity, (\lowercase{\romannumeral3}) an audio encoder $E_a$ for extracting the per-frame audio feature, and at last (\lowercase{\romannumeral4}) a StyleGAN-based decoder $G$ to generate images that exhibits the transferred style of the reference frames onto the target image.

\noindent\textbf{Encoder.}
The content encoder $E_c$ consists of $L$ layers, which are constructed using $L-2$ residual down-blocks along with two additional convolution blocks.
The encoder extracts the intermediate content features $f_c^i$ from each of the layers, $i\in {1, ..., L}$, later used for residual mapping to the generator $G$. The final content feature $f_c \in \mathbb{R}^{12 \times 12 \times 512}$ is used as an input of the decoder. 

The structure of the style encoder $E_s$ is similar to that of $E_c$. It is comprised of $L$ layers, each producing intermediate style features $f_s^i$, $i\in {1, ..., L}$. Note that $E_s$ has an additional fully-connected layer to yield a sparse feature $f_s \in \mathbb{R}^{512}$ of the reference image. 
We make the reference feature sparse, ensuring that the generator solely captures the broad attributes of the reference image while disregarding its finer structural details.

The audio encoder $E_a$ receives mel-spectrogram $A$ as input. We use the self-attentive pooling layer introduced in \cite{cai2018sap} to focus on important frame-level features. 
The final audio feature $f_a \in \mathbb{R}^{512}$ is concatenated with the style feature $f_s$ to formulate $f_\mathcal{W}=\{f_s, f_a\} \in \mathbb{R}^{1024}$ as the $\mathcal{W}$ space for the style mapping to the generator layers.

\noindent\textbf{Decoder.} The overall structure of the decoder follows the StyleGAN2~\cite{Karras2019stylegan2}, with $L$ hierarchical layers. With the content feature $f_c$ as input and  $f_\mathcal{W}$ to modulate the convolution kernel weights of the generator, the decoder generates faces dubbed with the guidance of style and audio features upon the target image. Previous one-shot audio-driven works that utilize direct skip connections~\cite{Parjwal2020wav2lip} have higher reliance to the structural information, like the poses and mouth shapes, of the reference image. That is, with a reference image with a dissimilar pose of the target image or ground truth mouth shape, the model struggles to generate high-fidelity results.
Instead, we employ style modulation to convey the identity information, which eventually helps the robustness of distinct poses and mouth shapes from the reference images. 
We present empirical results in Section \ref{sec:ablation} to demonstrate that style modulation of reference image is more effective than skip-connections or direct input injections for high-quality lip-sync generation.

\subsection{Design of Vision Transformer Blocks}

With the sparse feature of the reference image delivered to the decoder, it is insufficient to reconstruct high-fidelity details of the target identity. As a solution, we incorporate Vision Transformer (ViT)~\cite{dosovitskiy2020vit} to restore these intricate details. We adopted the attention mechanism of ViT to understand the meaningful patterns and relationships between global image patches. With the aid of global attention, our framework is able to focus on lip-oriented regions even for misaligned reference frames compared to the target. 

The ViT blocks are strategically designed to focus on the lip regions, which in our scenario corresponds to the lower half of the image.
ViT blocks are attached into the final two layers, namely the $L-1$ and $L$-th layers, of the decoder. We empirically found that attention in the final two layers were the most efficient and effective (see experimental results in the supplementary material).
The lower half of the intermediate feature of the decoder $f_d^{L-i}$, is used as the query ($Q$). The lower half of the intermediate style feature $f_s^i$ is used as the key ($K$) and value ($V$). 
We first compute the attention output with the following equation :
\begin{align} \label{eq:vit}
\text{Attention}(Q,K,V)=\text{softmax} \left( \frac{QK^\mathbf{T}}{\sqrt{d_k}} \right) \cdot V,
\end{align}
where $Q,K,V$ are each the output of layer normalization over corresponding features, and $d_k$ is the dimension of the key vector. The attention output is then added to the the intermediate features. We used eight multi-heads to concat the attention score, and linearly transformed them with multi-level perceptron (MLP) layers to produce the final attention layer.
Finally, The output of the ViT block is concatenated with the upper half of the intermediate feature $f_d^{L-i}$, then passed to the style modulation layer. 

The patch size of ViT block is empirically set to $\frac{H_i}{32}\times\frac{W_i}{32}$ for the $i$-th intermediate features layers. 
The ViT blocks consist of two attention layers, resulting in a total of four attention layers considering the entire architecture.
We name each of these layers $Att_{ij}$, where $i \in \{L-1, L\}$ and $j \in \{1,2\}$.

\input{fig/fig_main_ex}

\subsection{Loss Function}

In the training phase, we use five consecutive target frames aligned with the audio clip, while the reference frames are randomly chosen . We use the following training objectives to enhance image quality and synchronization accuracy.

\noindent\textbf{Reconstruction Loss.}
The reconstruction loss $L_{rec}$ is composed of an $L_1$ pixel loss and a perceptual loss:
\begin{align} \label{eq:rec_loss}
\mathcal{L}_{rec}= \| I_t - I_{out} \|_1 +  \sum_{i=1}^L \lambda_i \|  \phi_i(I_t) - \phi_i(I_{out}) \|_1,
\end{align}
where $\phi_i$ is the $i$-th layer of the VGG network and $L$ is the number of VGG layers. We use different weight $\lambda_i$ for each layer, increasing for deeper layers. 

\noindent\textbf{GAN Loss.}
To maintain high fidelity of the generated image, we use adversarial loss $\mathcal{L}_{adv}$ commonly used in generative networks: 
\begin{align} \label{eq:gan_loss}
\mathcal{L}_{adv} = \mathbb{E} \left[\log(1+\exp({D(I_{out}})) + \log(1+\exp({-D(I_{t}})) \right],
\end{align}
where $D$ is the StyleGAN2~\cite{Karras2019stylegan2} discriminator, trained jointly with our generator. 

\noindent\textbf{Sync Loss.}
Following \cite{Parjwal2020wav2lip}, we additionally train a gray-scale sync discriminator $S$, consisting of a vision encoder $S_v$ and audio encoder $S_a$. The encoder architecture follows \cite{chung2020voxtrainer} with self-attention pooling after ResNet layers. The details of our modified sync discriminator architecture can be found in the supplementary material.
The sync discriminator is trained with a binary-cross entropy loss (eq. \ref{eq:syncnet_loss}) to increase the cosine similarity (eq. \ref{eq:sync_sim}) of the vision and audio features of five consecutive frames that are in-sync ($y_i=1$), while pursuing the opposite for frames that are off-sync ($y_i=0$).  
\begin{align} \label{eq:sync_sim}
p_i(I, A) = \frac{S_v(I_{i-2:i+2})^\mathbf{T}\cdot S_a(A_{i-2:i+2})}{\|S_v(I_{i-2:i+2})\| \cdot \|S_a(A_{i-2:i+2})\|} ,
\end{align}

\begin{align} \label{eq:syncnet_loss}
\mathcal{L}_{sync} = -\mathbb{E} \left[ y_i \log(p_i) + (1-y_i)\log(1-p_i) \right].
\end{align}

During training the RADIO framework, we fix the weights of the pre-trained sync discriminator, and enhance the synchronization quality by the same sync-loss.

The overall training loss is formulated as follows:
\begin{align} \label{eq:total_loss}
\mathcal{L}_{total} = \mathcal{L}_{rec} + \lambda_{adv}\mathcal{L}_{adv} + \lambda_{s}\mathcal{L}_{sync},
\end{align}
where $\lambda_{adv}$ and $\lambda_{sync}$ are balancing weights.


%% file: fig/fig_main_ex.tex
\begin{figure*}[t!]
\begin{center}
\includegraphics[width=\textwidth]{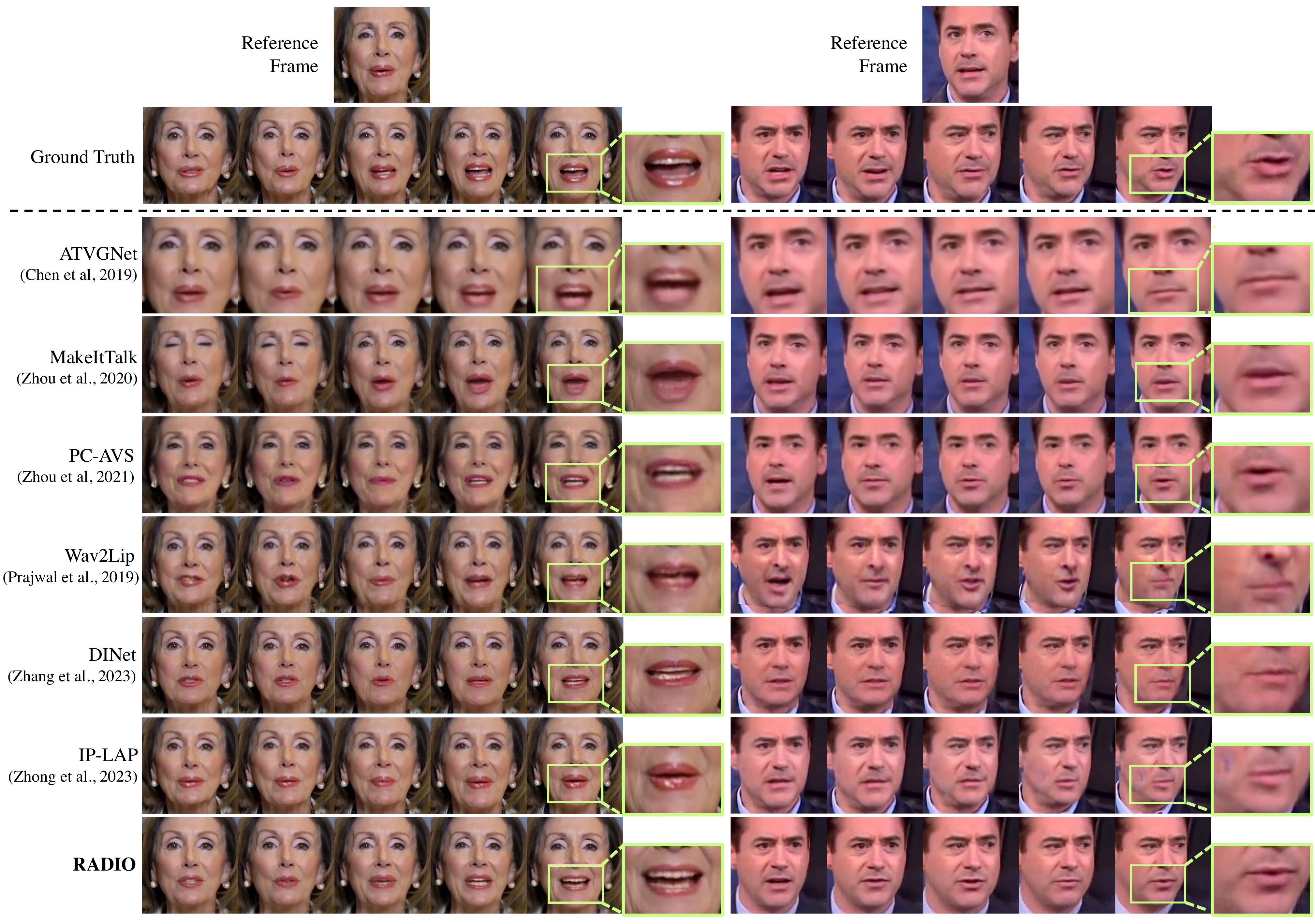}
\end{center}
\caption{
\textbf{Qualitative comparison with baselines.}
We visualized the dubbed results for five adjacent frames from the HDTF (left) and VoxCeleb2 (right) dataset, and zoomed-in images of the mouth region for the closer inspection. The reference frame for HDTF clip was a frontalized face with closed mouth, and the reference frame for VoxCeleb2 was a face facing the left side.  Our method showed the highest fidelity and accurately synced results, agnostic to the reference image. 
}
\label{fig:main_ex}
\end{figure*}

%% file: section/4_experiment.tex

\input{tab/tab_exp1}

\section{Experiments}

\subsection{Experimental Settings} \label{sec:data}

\noindent\textbf{Datasets and Preprocessing.} We trained our RADIO and sync-discriminator with LRW~\cite{chung2016lrw} dataset, a commonly used audio-visual dataset with 1,000 utterances of 500 different words. For evaluation, we used 50 randomly selected videos from HDTF~\cite{zhang2021hdtf} and VoxCeleb2~\cite{chung2018voxceleb2} datasets. The HDTF dataset primarily consists of videos with frontalized head poses, and VoxCeleb2 consists of videos with a wide range of head poses.  All videos are resampled to a frame rate of 25 FPS.

For the image preprocessing, we first detected faces and cropped the images with FFHQ-alignment~\cite{karras2019style}, then resized them to resolution $192 \times 192$. This alignment crop contains the whole lower half of faces,  with both the mouth and nose vertically positioned at the center.
We used $L=6$ layers for all encoders and decoder to match the resolution. 
For the audio preprocessing, we first converted audios to a sample rate of 16 kHz, then extracted mel-spectrograms using FFT window size 1,280, a hop length of 160, and 80 mel filter-banks.

\noindent\textbf{Baselines.}
We compared our methods with person-agnostic talking face methods, including recent methods that claim to be state-of-the-art. The baselines include \textbf{ATVGNet}~\cite{chen2019hierarchical}, \textbf{MakeItTalk}~\cite{zhou2020makeittalk}, \textbf{PC-AVS}~\cite{zhou2021pcavs}, \textbf{Wav2Lip}~\cite{Parjwal2020wav2lip}, \textbf{DINet}~\cite{zhang2023dinet}, and \textbf{IP-LAP}~\cite{zhong2023iplap}. Note that the first three methods generate the entire head driven by the audio, so the alignment of synthesized faces differs from the ground truth.
The last three and RADIO correspond to inpainting-based methods, with slightly different masked regions around the mouth.
DINet~\cite{zhang2023dinet} and IP-LAP~\cite{zhong2023iplap} authors used multiple reference frames for better grasp of identity. For a fair comparison, all baselines used only the first frame of the video as an identity reference. 



\subsection{Qualitative Evaluation} \label{sec:evaluation}

Figure \ref{fig:main_ex} shows the qualitative comparisons for an example clip of five adjacent frames.
Specifically, for HDTF~\cite{zhang2021hdtf} video clip on the left, generated images should resemble ground truth frames with widely opening mouths, given a source face image with a closed mouth. For VoxCeleb2~\cite{chung2018voxceleb2} video clip on the right, methods should generate realistic faces tilted rightwards, given a source face image facing the left.
Except for ATVGNet~\cite{chen2019hierarchical}, we aligned all generated frames using FFHQ alignment for clear comparison.

In comparison to other methods, our approach generated faces that closely resemble the ground truth in terms of visual fidelity and lip synchronization. Methods that synthesize the whole head \cite{chen2019hierarchical, zhou2020makeittalk, zhou2021pcavs} showed poor identity preservation and the alignment highly deviated from the ground truth due to the missing resource to drive the pose.
Wav2Lip~\cite{Parjwal2020wav2lip} generated blurry lower faces with indistinct mouth attributes. DINet~\cite{zhang2023dinet} and IP-LAP~\cite{zhong2023iplap} showed poor performance, with the generated lip regions more closely resembling the source image than the ground truth. IP-LAP failed to generate open lips throughout the whole video, and created artifacts for deviating pose alignments, \eg seventh row.
More qualitative comparisons for challenging scenarios are provided in the supplementary material.

\noindent\textbf{Experiments for Robustness.} 
In this section, we explore the robustness of our method by dubbing the same target frames with various reference images. Figure \ref{fig:robust} shows generated images of RADIO using three different reference images. Specifically, each reference image are facing the front (A), the right with a closed lip (B), and the left with an open lip (C). 
An ideal scenario is to consistently generate accurate lips regardless of the varying pose and mouth shapes. Remarkably, our method demonstrated almost zero sensitivity to the reference image, as evidenced by the dubbed results. The synthesized images also closely resembled the ground truth, with nearly identical mouth shapes. With this guarantee of robustness, RADIO can be used without the need for additional reference frame selection processes.

\subsection{Quantitative Evaluation}

We evaluated methods with the following metrics to measure the reconstruction and lip synchronization quality. \textbf{PSNR}, \textbf{MS-SSIM}~\cite{wang2003msssim} and \textbf{LPIPS}~\cite{zhang2018perceptual} measure the pixel-wise and feature-wise similarity between generated and ground truth images. 
The SyncNet~\cite{Chung2016syncnet}  confidence score \textbf{Sync-C} and distance score \textbf{Sync-D} measure the audio-visual synchronization quality. We used the officially released version of SyncNet\footnote{\url{https://github.com/joonson/syncnet_python}} \cite{Chung2016syncnet} for a fair comparison. 
Lip landmark distance (\textbf{LMD}) measures the normalized landmark distance of the mouth between generated and ground-truth images for lip synchronization evaluation.

Evaluating raw images with metrics designed for reconstruction quality would be unfair due to the varying sizes of the generation regions across different methods,
\eg, DINet~\cite{zhang2023dinet} synthesizes only the small region around the mouth using its own cropping algorithm, Wav2Lip~\cite{Parjwal2020wav2lip} inpaints the lower half of the tightly cropped face, and RADIO inpaints the lower half of the FFHQ aligned face which includes more background for generation. Since the final dubbed videos only need the tightly zoomed face to be attached, we employed a cropping method that zooms in on the faces with the same ratio and then resized them to the same resolution for evaluation.
Please refer to the supplementary material for details on the cropping method.

Table \ref{tab:main_exp1} shows the quantitative comparison between competing methods. Our proposed method achieved the best performance on all visual quality metrics (PSNR, MS-SSIM, LPIPS). While IP-LAP \cite{zhong2023iplap} generated comparable visual quality synthesizing talking face to ours, it fell short in audio-visual synchronization, such as Sync-C/D and LMD. The reason for this is that IP-LAP focuses on the personalized facial traits and only uses landmarks priors for lip synchronization, which can be quite inaccurate from a single image. Regarding audio-visual synchronization, our method performed the best on LMD. Although Wav2Lip \cite{Parjwal2020wav2lip} performed the best on Sync-C/D, this result may be attributed its use of SyncNet, which was pre-trained on the same LRS2~\cite{afouras2018lrs2} dataset as the official SyncNet used for our evaluation. PC-AVS \cite{zhou2021pcavs} also achieved high Sync-C/D scores, but generated jerky lip movements to match the audio, which degraded the LMD scores. While our method demonstrated the third-best performance on Sync-C/D, it's noteworthy that this score is the closest to the ground truth. RADIO generated the most synchronized natural lip movements, as supported by qualitative results. Compared to baselines, RADIO is the only method that can robustly deliver both fidelity preservation and synchronization.

\input{fig/fig_robust.tex}

\subsection{Ablation Study of ViT blocks}
\label{sec:ablation}

In this section, we first analyzed the ViT blocks via visualizing the attention map. Then, we demonstrated the effectiveness of our ViT design with a thorough ablation study.

\noindent\textbf{Attention  Visualization.} Figure \ref{fig:vit} displays the attention map, which highlights the important region in the reference image for each corresponding patch of the synthesized image.
Arbitrary identity is presented on the first column in each row. For each row, the patch location is illustrated with green on the generated images in the upper half, and the reference image with the corresponding attention map in the lower half.
In particular, we visualized the attention map of $Att_{5,2}$, which is located in the second layer of the attention block in the fifth decoder layer. 
The attention map of ViT block in the last decoder layer is also visualized in the supplementary material.

Even with differently synthesized mouth shapes compared to the reference image, patches near the mouth successfully attended to important features like cheeks and similar locations of the mouth, as seen in the second and third columns.
In contrast, patches unrelated to mouth details primarily focused their attention on corresponding regions on the reference face, with less attention directed towards the mouth, as evident in the fourth and fifth columns. This phenomenon showed that our proposed ViT block leverages its global context understanding and semantic knowledge to successfully focus on lip-oriented details. With the capacity of ViT blocks to substantially guide global attention, RADIO can generate high-fidelity talking faces, even with misaligned reference frames.

\input{fig/fig_vit}

\noindent\textbf{Ablation Study.} We further conducted ablation studies to validate the effectiveness of our proposed framework. In Table \ref{tab:ablation}, we quantitatively compared the performance while changing each component we proposed, evaluated on the LRW~\cite{chung2016lrw} validation dataset. 
We used PSNR and LPIPS to assess the perceptual quality compared to the ground truth. For assessing lip-sync accuracy, we calculated the similarity score between audio and visual SyncNet features (eq. \ref{eq:sync_sim}), employing our SyncNet pre-trained on the LRW train dataset. In this comparison, all models were trained for 210K iterations with a batch size of 16, with resolution scaled down to $96 \times 96$ for the sake of resource efficiency. 

We added up different parts of our model starting from a \textbf{baseline (A)}, which directly injects the reference frame information to the decoder consisting of no ViT blocks. 
In this configuration, the model takes concatenated target and reference frames as input to the visual encoder, following the input design of Wav2Lip~\cite{Parjwal2020wav2lip}. 
Instead of learning identity via style modulated convolution, the baseline uses the reference feature as input to the decoder, with decoder layers only modulated by the audio feature, \ie $f_\mathcal{W}=f_a \in \mathbb{R}^{512}$. Quantitative results show that the generated images become more susceptible to the influence of the reference frame, thereby compromising the synchronization quality.

\textbf{Method B} separates the reference and target frames using individual encoders, specifically the style encoder and content encoder. Then it utilizes the style feature $f_s$ to compose the latent space for style modulation, along with the audio feature, \ie $f_\mathcal{W}=\{f_a, f_s\} \in \mathbb{R}^{1024}$. In this regime, the model captures less structural information from the reference frame, which enables it to improve synchronization quality by reducing its dependence on the source mouth shape.
\textbf{Method C} builds upon method B by incorporating a fidelity mapping via a straightforward addition adding the lower half of reference frame features to the intermediate decoder layers, \ie, $f_d^{L-i} + f_s^i$. Even with this na\"ive skip-connection of reference frame, we observe improvements across perceptual evaluation metrics. However, this simple integration eventually degraded the synchronization quality, because of its sensitivity to the reference frame.

\textbf{Method D}, our RADIO framework, attached ViT blocks into the decoder layers to selectively extract high-fidelity details necessary for generating synchronized mouth movements. With the combination of modulated convolution via style features and fidelity mapping via ViT, our framework earned the most gain compared to the baseline model. This configuration demonstrated the best quantitative performance by simultaneously learning high-fidelity details and maintaining high synchronization.

\input{tab/tab_abl}

%% file: tab/tab_exp1.tex
\begin{table*}[t!]
\small
\centering
\scalebox{1}{
\begin{tabular}{l|ccccc|ccccc}
\toprule
& \multicolumn{5}{c|}{HDTF} & \multicolumn{5}{c}{VoxCeleb2}\\
Methods & \scriptsize{PSNR$\uparrow$} & \scriptsize{MS-SSIM$\uparrow$} &  \scriptsize{LPIPS$\downarrow$} & \scriptsize{Sync-C$\uparrow$/D$\downarrow$}  & \scriptsize{LMD$\downarrow$} &  \scriptsize{PSNR$\uparrow$} & \scriptsize{MS-SSIM$\uparrow$} & \scriptsize{LPIPS$\downarrow$} & \scriptsize{Sync-C$\uparrow$/D$\downarrow$}  & \scriptsize{LMD$\downarrow$} \\

\midrule
GT & 100.0 & 1.0 & 0.0 & 8.716*/6.853* & 0.0 & 100.0 & 1.0 & 0.0 & 6.363*/8.014* & 0.0 \\
\cdashlinelr{1-11}

ATVGNet \cite{chen2019hierarchical} & 30.101 & 0.780 & 0.170 & 6.422/8.640 & 9.893 & 29.555 & 0.716 & 0.194 & 5.203/9.087 & 11.454 \\
MakeItTalk \cite{zhou2020makeittalk} & 29.230 & 0.646 & 0.232 & 5.226/9.784 & 10.963 & 29.066 & 0.619 & 0.233 & 4.472/9.759 & 12.415 \\
PC-AVS \cite{zhou2021pcavs} & 29.667 & 0.730 & 0.173 & \underline{9.060}/\underline{6.461} & 10.305 & 28.997 & 0.687 & 0.210 & \underline{7.179}/\underline{7.413} & 11.773 \\
Wav2Lip \cite{Parjwal2020wav2lip} & 30.645 & 0.818 & 0.135 & \textbf{9.918}/\textbf{6.105} & 7.882 & 30.487 & 0.791 & 0.146 & \textbf{7.397}/\textbf{6.105} & \underline{7.913} \\
DINet \cite{zhang2023dinet} & 30.892 & 0.858 & 0.088 & 7.969/7.372 & \underline{7.632} & 30.077 & 0.766 & 0.145 & 5.466/8.682 & 8.781 \\
IP-LAP \cite{zhong2023iplap} & \underline{32.111} & \underline{0.902} & \underline{0.068} & 7.072/8.075 & 7.702 & \underline{31.395} & \underline{0.844} & \underline{0.106} & 4.942/8.765 & 8.589 \\
\textbf{RADIO} & \textbf{33.939} & \textbf{0.909} & \textbf{0.058} & 8.310*/7.038* & \textbf{7.235} & \textbf{32.804} & \textbf{0.896} & \textbf{0.073} &  6.671*/7.856* & \textbf{7.544} \\
\bottomrule
\end{tabular}
}
\caption{\textbf{Quantitative comparison of baselines.} We measured the perceptual and lip-sync quality for baselines in HDTF and VoxCeleb2 datasets.
We denote the best scores in \textbf{bold}, second-best \underline{underlined}, and the closest Sync-C/D scores to the ground truth with * mark.
}
\label{tab:main_exp1}
\end{table*}

%% file: fig/fig_robust.tex
\begin{figure}[t]
\begin{center}
\includegraphics[width=1\columnwidth]{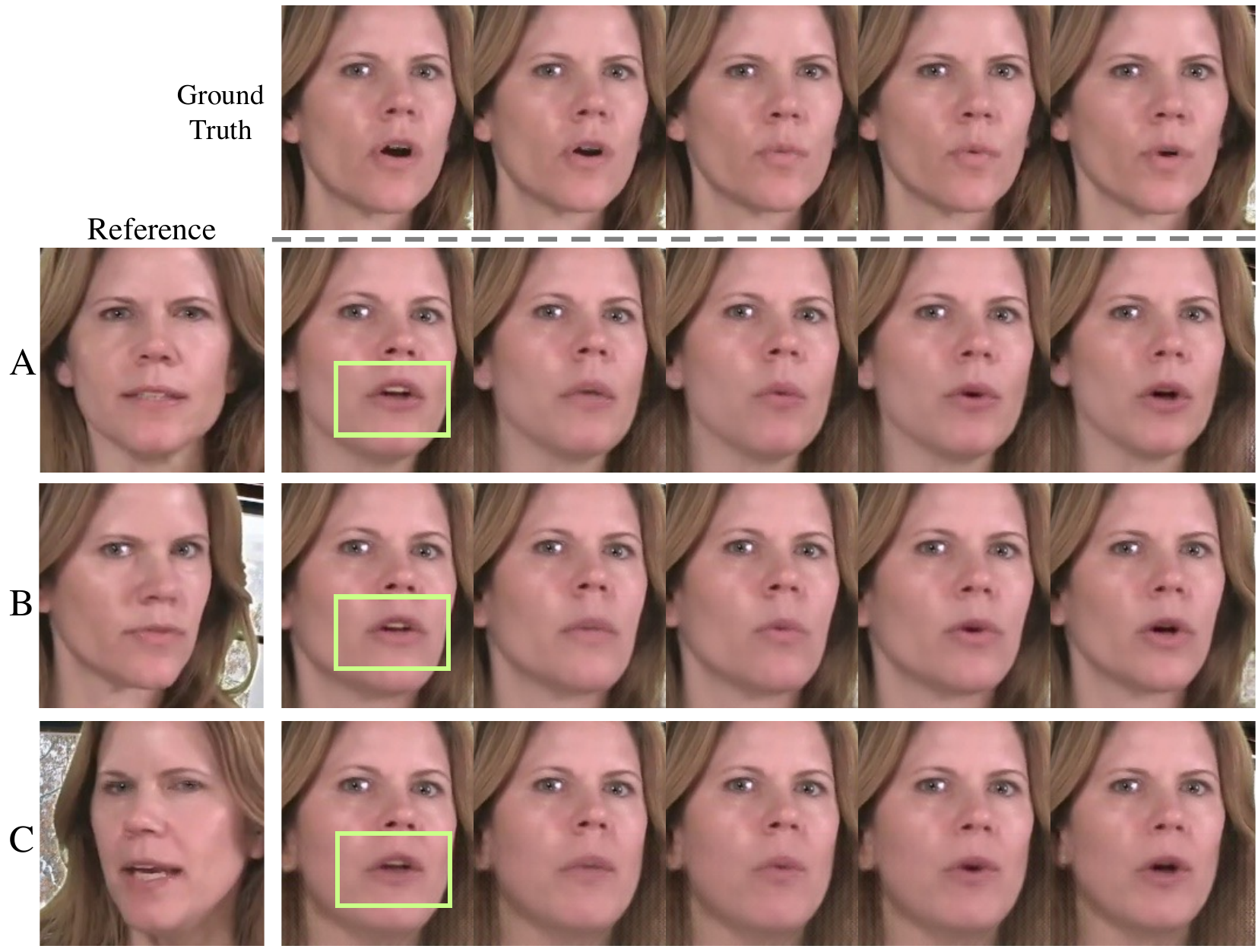}
\end{center}
\caption{
\textbf{Qualitative validation of the robustness of RADIO.} Our method consistently produced accurately dubbed videos, showcasing its robustness in generating lip-synchronized content regardless of the variations in the reference frames.
}
\label{fig:robust}
\end{figure}

%% file: fig/fig_vit.tex
\begin{figure}[t]
\begin{center}
\includegraphics[width=\columnwidth]{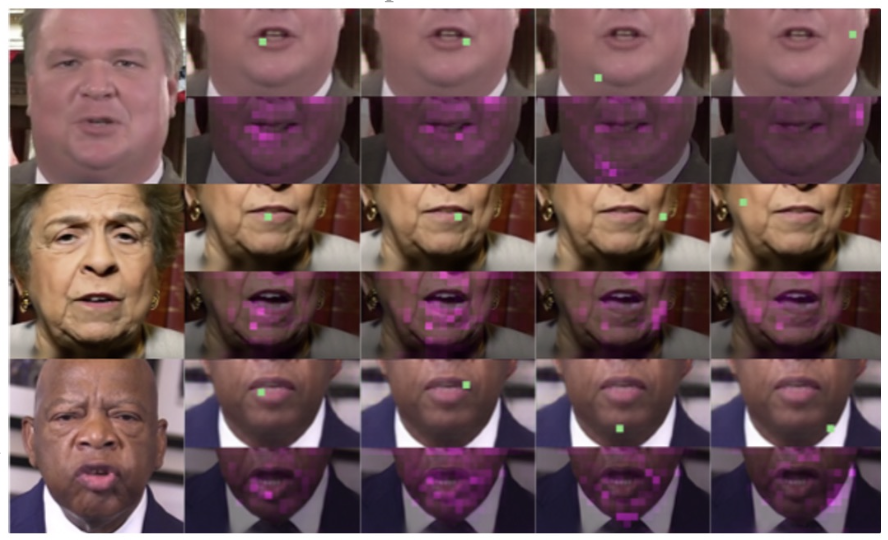}
\end{center}
\caption{
\textbf{Analysis of ViT Blocks.} For three different identities, we visualized the green patches on generated frames (upper half) with the attention map on reference frames (lower half). Our well-trained attention layer consistently focused on the globally relevant region of the reference frame for each local patch.
}
\label{fig:vit}
\end{figure}

%% file: tab/tab_abl.tex




\begin{table}[t]
\small
\centering
\scalebox{0.97}{
\begin{tabular}{@{}llccc@{}}
\toprule
\multicolumn{2}{l|}{Method}                                & \multicolumn{1}{c}{PSNR$\uparrow$} & \multicolumn{1}{c}{LPIPS$\downarrow$} & \multicolumn{1}{c}{Sync$\uparrow$} \\ \midrule
A & \multicolumn{1}{l|}{Baseline}  & 32.672 & 0.040 &  0.520 \\
\cdashlinelr{1-5}
B & \multicolumn{1}{l|}{+ Style modulation} &  33.089  &  0.072  &  0.576  \\
C & \multicolumn{1}{l|}{+ Fidelity mapping w/o ViT} & 34.493 & 0.037  & 0.554\\
\rowcolor[gray]{0.91}
D & \multicolumn{1}{l|}{+ Fidelity mapping w/ ViT} &  \textbf{34.938} & \textbf{0.031} & \textbf{0.609}\\ \bottomrule
\end{tabular}
}
\caption{\textbf{Ablation study with quantitative evaluation on LRW~\cite{chung2016lrw}.} We varied the latent space for modulated convolution and the method for fidelity mapping. 
Our framework (D) achieved the most improvement compared to the baseline.
}
\vspace{-3mm}
\label{tab:ablation}
\end{table}

%% file: section/5_conclusion.tex
\section{Conclusion}

In this paper, we presented an efficient framework that generates accurately dubbed faces with lip-oriented details preserved from a single source image.
Our work especially standed out in the challenging yet under-investigated scenario where the face orientation and lip shape between the source and target frames are significantly different. 
RADIO adapts to the identity of a person via StyleGAN2 style modulation, whilst reducing the reliance on facial alignment.
With the aid of ViT blocks, RADIO is finally able to synthesize faces with high-fidelity details by focusing on the important facial attributes of the reference image.
Through extensive experimentations, our method has demonstrated its unique capability to consistently generate high-fidelity videos while maintaining precise lip synchronization. This achievement establishes it as the new state-of-the-art in the field of one-shot audio-driven talking face generation. 
Considering the simplicity and practical applicability of our framework, we look forward to wide usage in future work.


%% file: section/6_acknowledgements.tex
\section{Acknowledgements}


This work was supported by the National Research Foundation of Korea (NRF) grant funded by the Korea government Ministry of Science and ICT (MSIT) (No. 2021R1G1A1094379), 
and in part by MSIT under the Information Technology Research Center (ITRC) support program (IITP-2023-RS-2023-00258649) supervised by the Institute for Information \& Communications Technology Planning \& Evaluation (IITP), 
and in part by the IITP grant funded by MSIT (Artificial Intelligence Innovation Hub) under Grant 2021-0-02068, 
and in part by IITP grant funded by MSIT (No.2020-0-01336, Artificial Intelligence Graduate School Program (UNIST), No.2022-0-00264, Comprehensive Video Understanding and Generation with Knowledge-based Deep Logic Neural Network), and NRF grant funded by MSIT (No. 2.220574.01).



%% file: section/X_appendix.tex
\section{Architectural Details} 
\label{sec:architectural_details}
In this section, we provide a detailed description of the RADIO encoders and the sync discriminator.

\textbf{Encoders.} The two encoders ($E_c$, $E_s$) mentioned in the main paper have a similar structure consisting of residual-down blocks. The $E_c$ and $E_s$ receive RGB frames with $192 \times 192$ resolution and pass four numbers of a residual-down block along with two additional convolution blocks. A single residual-down block involves two convolutional layers (kernel size=3) and LeakyReLU activation with a skip connection that adds intermediate features passed through the additional convolutional layer (kernel size=1). The $f_c \in \mathbb{R}^{12 \times 12 \times 512}$, which is a output of $E_c$, is fed to generator $G$. The only difference between $E_c$ and $E_s$ is that $E_s$ extracts $f_s \in \mathbb{R}^{1 \times 1 \times 512}$ via a spatial dimensional global average pooling operation and a fully-connected layer after four residual-down blocks.

The audio encoder $E_a$ receives mel-spectrogram as inputs and encodes to $f_a \in \mathbb{R}^{1 \times 1 \times 512}$ through 2D convolutional layers. The $E_a$ is implemented to follow~\cite{chung2020voxtrainer} structure.
\cite{chung2020voxtrainer} is considered one of the most effective architectures for speaker recognition using audio inputs and comprises SE layers and self-attention pooling with ResNet layers. 
We modified the activation function as LeakyReLU and normalization as instance normalization.

\textbf{Sync Discriminator.} The sync discriminator consists of an encoder that receives mel-spectrogram as input and an encoder that receives facial images as input. The audio encoder features follow exactly the structure of~\cite{chung2020voxtrainer}, while the visual encoder utilizes channel-attention and spatial-attention operations instead of self-attention of~\cite{chung2020voxtrainer}. 
This is because it is important to concentrate on the mouth's shape within the face image or the local area around it.
Finally, the sync discriminator is pre-trained with a loss function (eq. 4 in main paper)  to increase the cosine similarity (eq. 5 in main paper) of the vision and audio features so that we can provide superior audio-video synchronization errors during the RADIO training scheme.

We pre-trained the sync discriminator on the LRW~\cite{chung2016lrw} dataset using tightly zoomed face images with a resolution of $144 \times 144$. This approach allowed the discriminator to focus specifically on the lip shape of the synthesized facial image. In addition, the images were converted to grayscale, making the sync discriminator color-agnostic and enabling it to focus solely on learning the sync accuracy of mouth shapes.

\input{fig/fig_crop}

\section{Pre-processing algorithm for evaluation.}
Including our proposed method, baselines generated different sizes of images with different alignments and different target regions for synthesis. In order to evaluate quantitative metrics fairly, we applied a pre-processing algorithm to all generated images before comparison. First, we aligned all baseline methods with FFHQ alignment~\cite{Karras2019stylegan2}. Then, we applied a face cropping method based on the DINet~\cite{zhang2023dinet} masking algorithm. Last, we resized tightly cropped faces to the same resolution. 

Figure \ref{fig:crop} depicts the face cropping method. We assumed that the tip of the nose (the thirty-fourth point of the facial landmark) is in the center of the human face. Then, we computed two radius values: $Radius$ $H$, which measures the distance between the highest point and the thirtieth point along the y-axis of the facial landmark, and $Radius$ $W$, which measures the distance between the fifty-fifth point and the forty-ninth point along the x-axis of the facial landmark.  We then set the final $Radius$ value as the maximum value between these two distances. Finally, we cropped the attached facial image with this $Radius$ value, starting from the thirty-fourth point on the facial landmark.

\section{Baseline Models}
In this section, we describe additional details about baselines 
mentioned in Section 4.1 of the main paper.

\textbf{ATVGnet.} ATVGnet~\cite{chen2019hierarchical} proposes constructing high-level representation (facial landmarks) from the audio signal and generating talking head videos conditioned on the facial landmark. ATVGnet leverages the pixel-wise loss with attention mechanisms to ensure temporal consistency and utilizes a regression-based discriminator to generate accurate facial shapes and realistic-looking images in the training scheme. Finally, ATVGnet can only generate $128 \times 128$ resolution videos and cannot keep up with the head motion of the target frames.

\textbf{MakeItTalk.} MakeItTalk~\cite{zhou2020makeittalk} also proposes an audio-to-landmarks approach for controlling the motion of lips while determining the specifics of facial expressions and the rest of the talking-head dynamics from an audio signal. After that, MakeItTalk generates talking head animations ($256 \times 256$ resolution) with a single image (cartoon or natural human) and predicted landmarks using image-to-image translation. MakeItTalk animates talking head videos based on facial landmarks extracted from audio signals. However, these facial landmarks are too sparse to describe lip motion details and do not represent significant head motion.

\textbf{Wav2Lip.} To the best of our knowledge, Wav2Lip~\cite{Parjwal2020wav2lip} is the first approach to utilize a pre-trained Sync Discriminator in a training scheme and generate the lower half masked of the target frame. This method guarantees high audio-visual synchronization. However, it is highly dependent on the reference frame by feeding with concatenating the reference frame and masked input frame, and generates blurry results.

\textbf{DINet.} DINet~\cite{zhang2023dinet} proposes a deformation inpainting network, which performs spatial deformation on feature maps of reference images to synthesize high-fidelity dubbing videos. DINet uses five reference facial images to create deformed features in order to align head poses and driving audio to preserve high-frequency details. In addition, DINet develops its masking algorithm around the lip, resulting in efficient inpainting synthesis of mouth shapes. Finally, DINet can generate high-resolution ($416 \times 320$) videos. Although DINet utilizes multiple reference images, the synthesized results vary sensitively depending on the selected reference images. Also, their framework is restricted to frontalized head poses, and generates artifacts when the mouth region covers the background.

\textbf{IP-LAP.} IP-LAP~\cite{zhong2023iplap} follows a two-stage training scheme like any other method of utilizing facial landmarks. IP-LAP is implemented to leverage facial sketch maps rather than face landmark coordinates so the framework can learn the driving face shapes clearly. Finally, IP-LAP aligns the twenty-five reference images using a warping-based alignment module and utilizes them to generate $128 \times 128$ resolutions while preserving the target head pose and expression. Despite the usage of a large number of reference images, the accuracy of the lip shape is insufficient to learn the audio-visual synchronization only with facial landmarks.

\section{Ablation Study of ViT design}

\input{fig/fig_vit_last}
\input{tab/tab_abl_vit}

\textbf{Attention Visualization of last ViT block.} Figure \ref{fig:vit_last} visualizes the attention map of $Att_{6,2}$, located in the second layer of the attention block within the last ($L=6$) decoder layer. The upper half of the figure displays the green patches on the generated frames, while the lower half presents the corresponding attention map on the reference image. In contrast to Figure 5 in the main paper, each patch attended to the entire image for the last ViT block. We relate this phenomenon to the hierarchical nature of StyleGAN2~\cite{Karras2019stylegan2}, which generates course-to-fine information for low-to-higher layers.
While the intermediate attention layer, \ie, $Att_{5,2}$, focused on the globally relevant features for each local patch, the last attention layer, \ie, $Att_{5,2}$, captured the fine-grained textures and colors across the entire image.

\textbf{Design of ViT attention layers.} 
We additionally present quantitative results, evaluated on the LRW~\cite{chung2016lrw} validation dataset, for an ablation study of RADIO with varying numbers of ViT layers. Our evaluation metrics include PSNR, LPIPS, and the similarity score between audio and visual features. To obtain the similarity score, both the audio and visual features are encoded using the encoders of our sync discriminator described in Section \ref{sec:architectural_details}. We specifically used our pre-trained sync discriminator, which was trained with the LRW~\cite{chung2016lrw} training dataset, for accurate evaluation.
For this experiment, all models were trained for 210K iterations with a batch size of 16, with resolution scaled down to $96 \times 96$, like the main ablation experiment. We only conducted the experiment with ViT on the last two layers of the decoder, because patches for earlier layers were too small to deliver semantically interpretable results. For example,  the feature resolution is $12 \times 12$ for the fourth decoder layer, which is too small to divide into patches.

In Table \ref{tab:attn_nlayer}, the baseline refers to the framework that generates audio-driven images by decoder layers modulated with style features, without additional components for fidelity mapping (method B in Table. 2 of the main paper).
We denote ViT$(n,m)$ as our RADIO framework with $n$ ViT blocks, consisting of $m$ attention layers. Note that in our main experiment, we applied ViT block to the last two decoder layers, with each block comprising two attention layers, \ie, ViT$(2,2)$. The results indicated that having two attention layers in a single ViT block was better than using only one layer. Additionally, employing ViT blocks in two decoder layers was more effective than placing them in a single decoder layer. Finally, ViT$(2,2)$ achieved the best PSNR, LPIPS, and sync similarity scores compared to the baseline.

\section{Additional Experimental Results}

In this section, we show the additional experimental results of RADIO in Figure \ref{fig:main_appendix12} and Figure \ref{fig:main_appendix}. Throughout all examples, our results consistently generated the most natural and realistic mouth shapes, with high synchronization accuracy compared to the ground truth. ATVGNet, MakeItTalk, and PC-AVS commonly failed to generate identity-preserving details. Wav2Lip consistently created blurry images and generated artifacts for extreme face poses. 
Especially in harsh scenarios, IP-LAP and DINet struggled to generate realistic-looking mouth shapes, due to the significant distortion caused by warping and deformation. The mouth shapes generated by these methods were similar across all time steps, which also led to a degradation in synchronization quality. Especially, DINET failed to generate realistic faces with extreme poses, as their framework is limited to generate frontalized faces.

\section{Limitation and Broader Societal Impact}

While our model excels in producing high-quality images around the mouth region, it struggles to generate a natural-looking background. 
During our evaluation, we observed that frames significantly misaligned with the reference frame exhibited artifacts in the background. 
This limitation is observed across all baseline models \cite{Parjwal2020wav2lip, zhong2023iplap}, but is more conspicuous for ours due to the alignment method that includes a larger portion of the background for generation. This issue can be easily fixed by borrowing a face-parsing model~\cite{yang2021gpen} to attach only the face region to the original video, thus improving the overall video quality.

Previous one-shot audio-driven frameworks have struggled to consistently generate realistic, high-fidelity frames. These challenges arise because they heavily rely on the reference image, which typically requires a frontalized pose with a neutral facial expression. 
In contrast, our reference-agnostic framework demonstrates exceptional capabilities in generating high-quality dubbed videos, even in the most challenging scenarios. This makes it suitable for a wide range of real-world industrial applications where diverse poses and expressions are encountered. We look forward to the application of our framework to generate realistic audio-driven faces for unseen speakers in real-time. Looking ahead, we aspire to enhance and extend our RADIO framework to support higher resolutions in the near future.

\input{fig/fig_main_appendix12}

\input{fig/fig_main_appendix3}

%% file: fig/fig_crop.tex
\begin{figure}[t]
\begin{center}
\includegraphics[width=0.95\columnwidth]{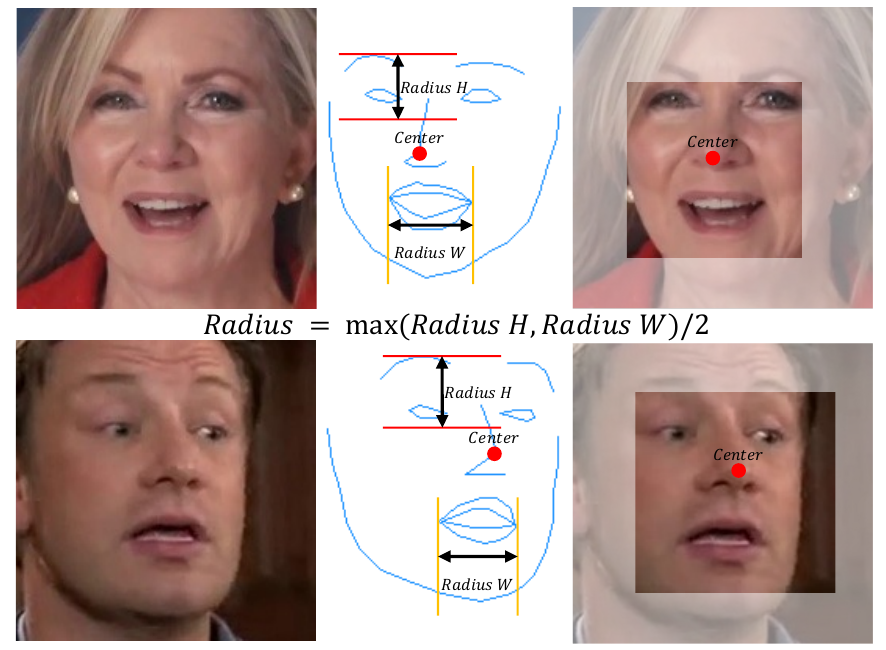}
\end{center}
\caption{
\textbf{Cropping method used for evaluation.} 
We applied this cropping method to all generated baseline results to evaluate the quantitative metrics with the same ground truth images.}
\label{fig:crop}
\end{figure}

%% file: fig/fig_vit_last.tex
\begin{figure}[t]
\begin{center}
\includegraphics[width=\columnwidth]{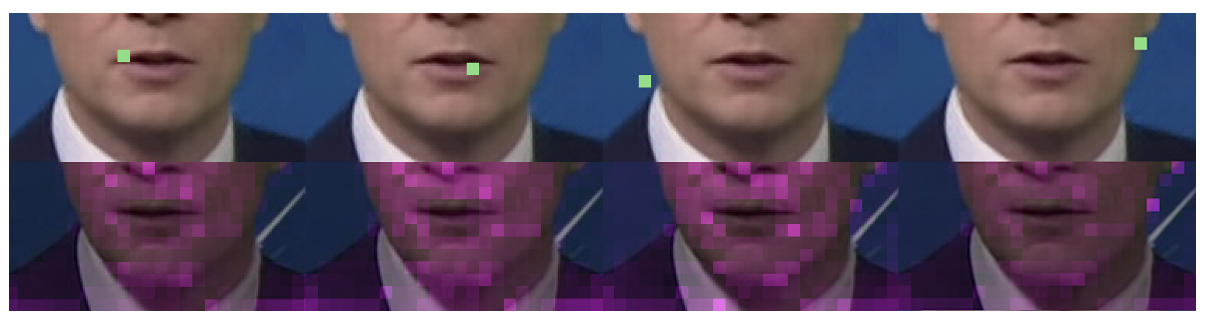}
\end{center}
\caption{
\textbf{Visualization of attention map in the last ViT block.} We visualized the green patches on generated frames (upper half) with the attention map on reference frames (lower half). 
}
\label{fig:vit_last}
\end{figure}

%% file: tab/tab_abl_vit.tex
\begin{table}[t]
\small
\centering
\scalebox{1}{
\begin{tabular}{@{}l|ccc@{}}
\toprule
Configuration  & PSNR$\uparrow$ & LPIPS$\downarrow$ & Sync$\uparrow$ \\ \midrule
Baseline &  33.089  &  0.072  &  0.576  \\
\cdashlinelr{1-4}
+ ViT$(1,1)$ &  34.637 & 0.033 & 0.557 \\ 
+ ViT$(1,2)$ &  34.757 & 0.032 & 0.559 \\ 
\rowcolor[gray]{0.91}
+ ViT$(2,2)$ &  \textbf{34.938} & \textbf{0.031} & \textbf{0.609} \\ \bottomrule
\end{tabular}
}
\caption{\textbf{Ablation for the different number of attention layers.} 
}
\label{tab:attn_nlayer}
\end{table}

%% file: fig/fig_main_appendix12.tex
\begin{figure*}[t]
\begin{center}
\includegraphics[width=0.88\textwidth]{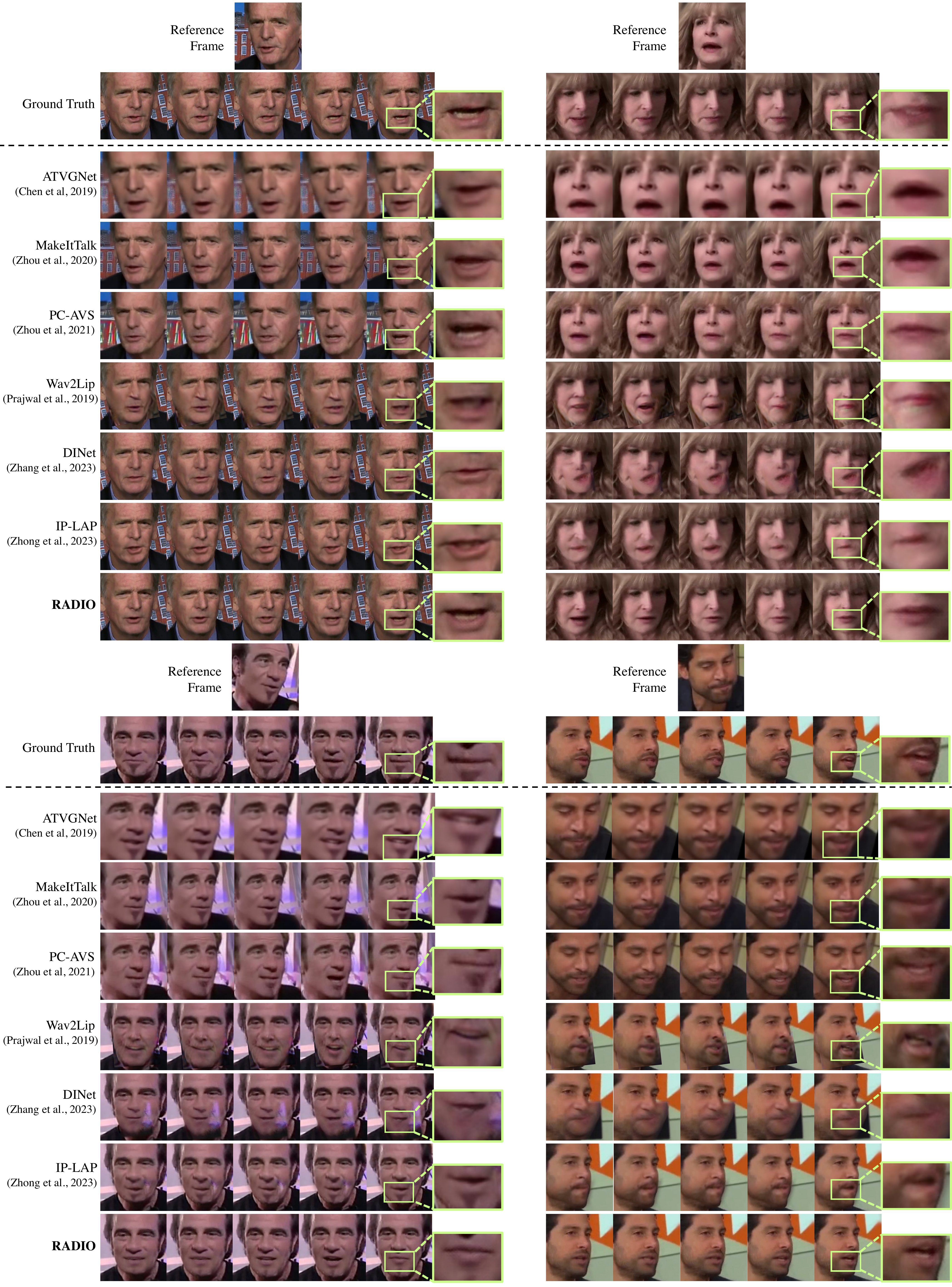}
\end{center}
\caption{
\textbf{Qualitative comparison with baselines.} We visualized the dubbed results for challenging scenarios where the ground truth pose and expression significantly differ from the reference frame.}
\label{fig:main_appendix12}
\end{figure*}

%% file: fig/fig_main_appendix3.tex
\begin{figure*}[t]
\begin{center}
\includegraphics[width=0.9\textwidth]{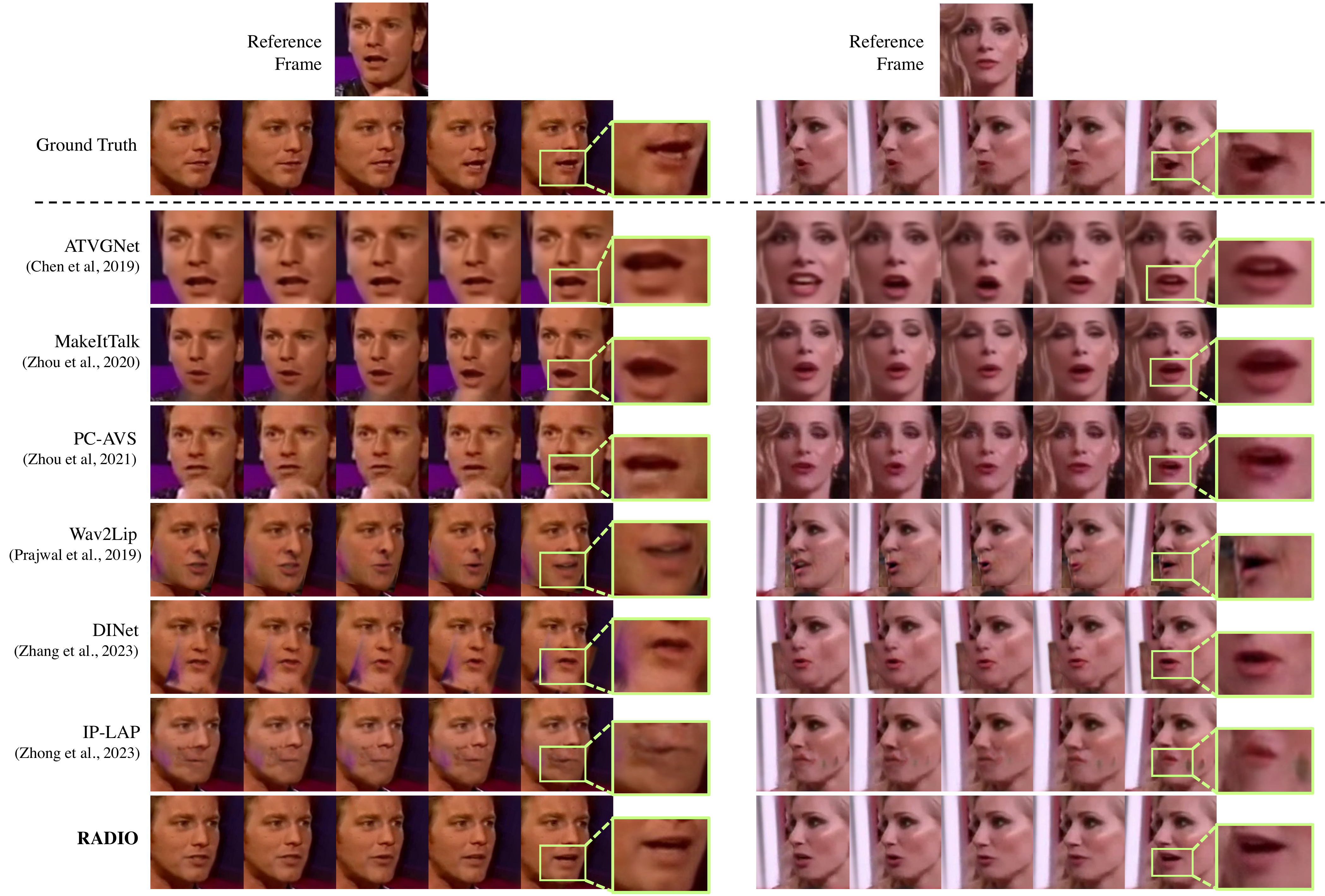}
\end{center}
\caption{
\textbf{Qualitative comparison with baselines.} We visualized the dubbed results for challenging scenarios
where the ground truth pose and expression significantly differ from the reference frame.}
\label{fig:main_appendix}
\end{figure*}